 \documentclass[final,1p,times]{elsarticle}

\usepackage{graphicx}
\usepackage{amssymb}
\usepackage{amsthm}
\usepackage{algpseudocode,hyperref,multirow}
\usepackage{amsmath, amscd,  mathtools,bm}

\usepackage[all]{xy}
\newcommand{\mb}[1]{\mathbf{#1}}
\newcommand{\md}[1]{\mathbb{#1}}

\newcommand{\thickhline}{%
	\noalign {\ifnum 0=`}\fi \hrule height 1pt
	\futurelet \reserved@a \@xhline
}

\theoremstyle{definition}
\newtheorem{theo}{Theorem}

\theoremstyle{definition}
\newtheorem{defi}{Definition}

\DeclareMathOperator\supp{Supp}

\DeclareMathOperator\Span{Span}
\usepackage{natbib}
\setcitestyle{square}

\def\betavec{{\boldsymbol\beta}}
\def\epsilonvec{{\boldsymbol\epsilon}}

\def\x{\mathbf{x}}
\def\xt{\tilde{\x}}

\def\xUnlab1{{\xt^{1}}}

\usepackage{subcaption}
\usepackage{algorithm}
\usepackage{ctable}
\usepackage{lineno}

\journal{Operation Research Letters}

\begin{document}

\begin{frontmatter}


\title{Scalable  Holistic  Linear Regression}



\author[label1]{Dimitris Bertsimas}

\address[label1]{Sloan School of Management and Operations Research Center, Massachusetts Institute of Technology, Cambridge, MA 02139}

\author[label2]{Michael Lingzhi Li}

\address[label2]{Operations Research Center, Massachusetts Institute of Technology, Cambridge, MA 02139}

\begin{abstract}
We propose a new scalable algorithm for holistic linear  regression building on Bertsimas \& King (2016). Specifically, we develop new theory to model significance and multicollinearity as lazy constraints rather than checking the conditions iteratively. The resulting algorithm scales 
	with  the number of samples $n$ in the 10,000s, compared to the low 100s in the previous framework. Computational results on real and synthetic datasets show it greatly improves from previous algorithms in accuracy, false detection rate, computational time and scalability.
\end{abstract}


\end{frontmatter}

%


\section{Introduction\label{sec:Introduction}}

In  this paper, we continue the research program initiated in  \cite{SLR1} to develop an algorithmic approach for holistic linear regression
in which we impose desirable properties simultaneously and a priori. 
 Using mixed integer optimization (MIO)  the earlier proposal   modeled  sparsity, pairwise collinearity and group sparsity using explicit constraints  but accounted 
 for significance and multicollinearity through a cutting plane and bootstrap approach.  
 The difficulty of using the cutting plane method is that it often  requires a large number of iterations to ensure that the model has appropriate significance and does not exhibit multicollinearity. This results in an algorithm that does  not scale  beyond  $n$, the number of samples,  in the low 100s when accounting   for significance and  multicollinearity.

Our goal in this paper is to propose a new scalable algorithm for holistic linear  regression. 
We propose a new way to impose significance and multicollinearity constraints explicitly that scales with $n$ in the 10,000s. This allows us to build linear regression models much more effectively and accurately than in earlier works. 

In our view, scalable holistic regression is important at it allows  linear regression models  to have  interpretability, robustness, 
significance and accuracy in a  systematic way. In contrast, today  the practice of regression is more of an art than science. Continuing on the vision in  \cite{SLR1}, the paper aspires 
to scale holistic regression further and make these methods easier to use in much larger problems.

The standard methodology for imposing significance  in linear regression is to use the Student $t$-statistic. However, the test is carried out after the linear regression model has been calculated, and does not optimally select a subset of covariates that are significant a priori. In the  \cite{SLR1} framework, summarized in Section \ref{sec:review}, significance is imposed iteratively leading to a cutting plane algorithm. 
\cite{car2017} explored significance of coefficients by adding heuristic constraints to set lower bounds on the coefficients. \cite{chung2017} used lazy constraints to ensure exact significance tests while deriving theoretical bounds for minimum power. In contrast to \cite{car2017} and similar to \cite{chung2017}, we use lazy constraints to ensure minimum power. However, instead of using the $t$-statistic, we appeal to the asymptotic normality results instead.

For multicollinearity, in a landmark paper  \cite{MC1} comprehensively reviewed the problem and concluded that there is no accepted way of dealing with this problem, citing ``there is a lack of attention for this problem in the statistics community.'' Various methods employed include principal component analysis to select the top $k$ variables to avoid multicollinear combinations, and variance inflation factors \cite{VIF} that provide a numerical quantity to determine how much the variance of a coefficient has been increased due to correlation with other variables. \cite{tamura2016,tamura2017} explored incorporating multicollinearity constraints using variance inflation factors (VIFs) and condition numbers (CNs) respectively. However, both of these concepts are only approximations of true multicollinear relationships. It is true that multicollinear relationships are sufficient for high VIF and CN, but they are not necessary, as shown in \cite{o2007} and \cite{lazaridis2007}, respectively. That means constraining on VIF or CN would potentially produce extra constraints that are not needed for solving multicollinearity. In this paper, we introduce new theory that provides both necessary and sufficient guarantees in relation to detecting multicollinearity.

\noindent
Specifically, our contributions in  this paper are as follows: 
\begin{enumerate}
    \item We continue the program in \cite{SLR1} and extend the formulation with significance constraints a priori.
	\item We develop a new theory of detecting multicollinearity by connecting multicollinearity to the eigenvectors of the design matrix $\bm{X}^T\bm{X}$, where $\bm{X}$ is the $n \times p$ matrix of the given data and use it   to impose multicollinearity constraints within an MIO   framework
	\item We present computational results on real and synthetic datasets that suggest the overall algorithm  for holistic regression scales with $n$ in the 10,000s, while the 
method in  \cite{SLR1} scales with $n$ in the low  100s when accounting for significance and multicollinearity.
\end{enumerate}
The structure of the paper is as follows.
In Section 2, we review the  work in \cite{SLR1}
on constructing a holistic framework for linear regression.
In Section 3, we introduce the $t$-statistic formulation to model significance.
In Section 4, 
we introduce a new formulation to model multicollinearity and  present  computational results  with synthetic and real-world data
that show the effectiveness of the method.
In Section 5,  we combine our proposals  with the framework introduced in   \cite{SLR1}, and  compare its 
 performance with recent models in the literature using real-world data.

\section{The Framework of   \texorpdfstring{\cite{SLR1}}{Bertsimas and King (2016)}}
\label{sec:review}
Given data $(x_i,y_i)$, $i=1,\cdots,n$, $x_i \in \md{R}^p$, $y_i \in \md{R}$,  \cite{SLR1} propose the following MIO:
\begin{eqnarray}
\min_{\betavec,\bm{z}}  & \displaystyle     \frac{1}{2}\|\mb{y}-\bm{X}\betavec\|^2 +\Gamma \|\betavec\|_1   \label{E0} \\
{\rm subject ~to:} 
&  -Mz_i\leq \beta_i\leq Mz_i , \qquad  i=1,\cdots ,p   \label{E1} \\
&\displaystyle    \sum_{i=1}^p z_i \leq k   \label{E2}  \\
& z_i=z_j , \quad   \forall i,j \in GS_m, ~\forall m \label{E3}  \\
&z_i+z_j\leq 1,\quad   \forall  i,j \in HC  \label{E4}  \\
& z_i \in \{0,1\} ,\quad i=1,\cdots, p   \nonumber
\end{eqnarray}
The term $\Gamma \|\betavec\|_1$ in the objective function (\ref{E0}) models robustness as seen in \cite{Robust}, who established the equivalence of the $\ell_1$ penalization and robustness.

 Constraints (\ref{E1}) and (\ref{E2}) model sparsity using the Big-$M$ framework that at most $k$ out of the $p$ variables are selected in the linear regression model. In this paper, the specification of $M$ follows from that in \cite{SLR1}.  Constraint (\ref{E3}) models group sparsity, i.e., variables in the set $GS_m$ are either all selected or none is selected. Finally, pairwise collinearity is modeled in Constraint (\ref{E4}) where HC is the set \[HC=\{(i,j): {\rm |Corr}(x_i,x_j)| \geq \rho\}\] for some predefined correlation $\rho$ cutoff. 

 \cite{SLR1} apply  the following iterative process
   to include constraints for significance and multicollinearity:
\begin{enumerate}
	\item Solve the MIO  (\ref{E0})-(\ref{E4}) to obtain a subset $S$ of  the coefficients $\{\beta_1,\cdots, \beta_p\}$.
	\item For  the set  $S$ the algorithm computes the significance levels for each of the variables via bootstrap methods, and calculates the condition number of the model. If a set $S$ produces undesirable results -- a condition number higher than desired, or a model with insignificant variables -- the algorithm generates the 
	 constraint
	 \[\sum_{i \in S} z_i \leq |S|-1\]
	  to exclude the set $S$ from consideration. The algorithm adds the constraint to Problem (\ref{E0}) 
	  and repeats the process until no such set $S$ is found. 
\end{enumerate}

\cite{SLR1} report computational results that demonstrate that  Model (\ref{E0}) is effective to solve problems up to   $n,p$ in the 1000s. However, when we include significance and multicollinearity constraints in a cutting plane methodology, the method scales up to $n,p$ in the low  100s and some times no solution is found after considerable computation time. 

\section{Imposing Significance Constraints}
Variable significance has long been one of the most important elements  in linear regression, and has served as a   proxy for variable selection and  causality studies. 

We first restate a standard result about the asymptotic guarantee of the normality of the least squares estimate of $\betavec$ to serve as the basis of our approach.
For a linear regression problem:
\[\mb{Y}=\mb{X}\betavec +\epsilonvec. \]
We have the following theorem, as proven in \cite{eicker1963asymptotic}:
\begin{theo}
    If $\epsilonvec$ is iid with $\md{E}[\epsilon_i]=0$ and $\md{E}[\epsilon_i^2]=\sigma^2$ for all $i$, and $\lim_{n \to \infty} \frac{\mb{X}^T\mb{X}}{n}=\mb{Q}$ is invertible, then we have:
    \[\frac{\sqrt{n}(\hat{\betavec}-\betavec)}{\sigma \sqrt{\bm{Q}}}\xrightarrow{d} N(0,1)\]
    Where $\hat{\betavec}$ is the least squares estimate of $\betavec$ with 
$$\hat{\betavec}=(\bm{X}^T\bm{X})^{-1}\bm{X}^T\bm{Y}.$$ 
\end{theo}
Note that normality is not part of the assumption here - in contrast to the $t$-test statistics used in \cite{chung2017}. Therefore, using such asymptotic results, we would assume when $n$ large enough, we have that:
\begin{equation}
\frac{\hat{\beta}_j-\beta_j}{\tilde{\sigma}\sqrt{\bm{K}^{-1}_{jj}}} \sim N(0,1).
\label{n-stat}
\end{equation}
Where  $\bm{K}=(\bm{X}^T\bm{X})^{-1}$
and $$\tilde{\sigma}=\sqrt{\frac{\mb{Y}^T(\bm{I}_n-\bm{X}(\bm{X}^T\bm{X})^{-1}\bm{X}^T)\mb{Y}}{n-p}}$$ is the least squares estimate of standard deviation $\sigma$. 
\subsection{Constructing Significance Constraints}
For a test of size $\alpha$, we first define the quantity $N_{sign}=\Phi^{-1}(1-\frac{\alpha}{2})$, the inverse cdf of the $N(0,1)$ distribution at point $1-\frac{\alpha}{2}$. Then,
we can impose the  normality test by requiring:
$$
\frac{|\beta_j|}{\tilde{\sigma}\sqrt{(\bm{X}_{\bm{z}}^T\bm{X}_{\bm{z}})^{-1}_{jj}}} \geq  N_{sign}z_j,
$$
Where $\bm{X}_{\bm{z}}$ is the model matrix constrained to the columns where $z_i=1$. This is equivalent to  the big $M$-constraints:
\begin{eqnarray}
	 \frac{\beta_j}{\tilde{\sigma}\sqrt{(\bm{X}_{\bm{z}}^T\bm{X}_{\bm{z}})^{-1}_{jj}}}+Mb_j &\geq  N_{sign}z_j \label{t-new} \\
	 -\frac{\beta_j}{\tilde{\sigma}\sqrt{(\bm{X}_{\bm{z}}^T\bm{X}_{\bm{z}})^{-1}_{jj}}}+M(1-b_j) &\geq  N_{sign}z_j   \label{t-new1}\\
	 b_j\in\{0,1\} , & j=1,\ldots,p, \nonumber
\end{eqnarray}
where $M$ is a large constant. These two constraints are used to model significance of level $\alpha$ without the need of the bootstrap. As the model matrix $\bm{X}_{\bm{z}}$ changes with the selection of $\bm{z}$, in implementation these constraints are implemented as lazy constraints to only be enforced when a feasible integer solution is reached, in a similar fashion to \cite{chung2017}.

In interest of brevity, we defer computational experiments and present results when combined with the multicollinearity detection as illustrated below.
\section{Multicollinearity Detection}
Given data $X$, we would like the design matrix to be free of multicollinear relationships so that $\det(\bm{X}^T\bm{X})$ is not very close to 0. We denote the columns of $\bm{X}$ as $\mb{X}_j$, $j=1,\cdots,p$. 

We introduce the vector $(1,\cdots,1)^T$ into the design matrix as a new column (the intercept)  and we can define the multicollinear relationship as:
\begin{defi}
	A set of variables $\mb{X}_1,\cdots \mb{X}_p$ has an \emph{$\epsilon$-multicollinear relationship} if for some $\mb{a} \in \md{R}^p$, $\|\mb{a}\|=1$, we have that:
	\begin{equation}
	\left\|\sum_{j=1}^n a_j\mb{X}_j\right\|<\epsilon .\
	\label{def1}
	\end{equation}
\end{defi}
The structure of this section is as follows:
\begin{enumerate}
	\item We first establish the key result   that connects the existence of an $\epsilon$-multicollinear relationship  (\ref{def1})
	 to the existence of an eigenvector $\mb{v}$ for the matrix $\bm{X}^T\bm{X}$ that has a small ($O(\sqrt{\epsilon})$) eigenvalue.
	\item Using the previous key result, we  find   multicollinear relations $(\mb{a}=(a_1,\cdots,a_p))$ using information from the small eigenvalues of the matrix $\bm{X}^T\bm{X}$.
	We introduce the  idea of a \emph{minimum-support multicollinear relationship}.
	\item We propose an algorithm that uses the theory from the previous steps to identify all the multicollinear relationships.
\end{enumerate}
\subsection{Key Result}
In this section, we establish a connection between the existence of an $\epsilon$-multicollinear relationship and the existence of a eigenvector $\mb{v}$  for the matrix $\bm{X}^T\bm{X}$ with a small ($O(\sqrt{\epsilon})$) eigenvalue:
\begin{theo} \label{theo2}
	Let $V=\{\mb{v}_1,\ldots ,\mb{v}_m\}$ be the set of orthonormal eigenvectors of $\bm{X}^T\bm{X} \in \md{R}^{p\times p}$ such that 
	the eigenvalues associated with  $V$ are less than  $\epsilon$. Then for $\mb{a}\in \md{R}^p$, $\|\mb{a}\|=1$:
	\begin{enumerate}
		\item[(a)]
		If $\left \|\sum_{j=1}^p a_j\mb{X}_j\right\|<\epsilon$, then there exists a vector $\mb{b} \in \md{R}^p$, $\|\mb{b}\|<(p-m)\sqrt{\epsilon}$ such that $\mb{a}-\mb{b} \in \Span(V)$.
		\item[(b)]
		If there exists a vector $\mb{b} \in \md{R}^p, \|\mb{b}\|<\sqrt{\epsilon}$ such that $\mb{a}-\mb{b} \in \Span(V)$, then we have:
		\[\left\|\sum_{j=1}^p a_j\mb{X}_j \right\|<\sqrt{(1+\lambda_{m+1}+\ldots+\lambda_p)\epsilon},\]
		where $\lambda_{m+1},\ldots,\lambda_p$ are the eigenvalues associated with the set of orthonormal eigenvectors of $M$ that have value greater or equal to $\epsilon$.	
	\end{enumerate}
\end{theo}
Theorem \ref{theo2} represents a weak equivalence between a small multicollinear relationship and the existence of a vector $\mb{a}$ that is close to $\Span(V)$, in the sense that there exists a small vector $\mb{b}$ with $\|\mb{b}\|<O(\sqrt{\epsilon})$ such that $\mb{a}-\mb{b}\in \Span(V)$. The proof is as follows:
\begin{proof}\leavevmode
	\begin{itemize}
		\item[(a)]  If  $m=p$, then every $\mb{a} \in \Span(V)$. Thus, we assume $m<p$ and  prove part (a) by contradiction.
		We assume there exists no $\mb{b} \in \md{R}^p$ with $\|\mb{b}\|<(p-m)\sqrt{\epsilon}$ such that $\mb{a}-\mb{b} \in \Span(V)$. 
		Let $\lambda_1,\ldots , \lambda_p$ be the corresponding eigenvalues to eigenvectors $\mb{v}_1,\ldots , \mb{v}_p$. 
		Note that we have $0\leq \lambda_1,\ldots , \lambda_m<\epsilon$, and $\epsilon\leq\lambda_{m+1},\ldots, \lambda_p$.
		We   write $\mb{a}$ as:
		\[\mb{a}=\alpha_1\mb{v}_1+\cdots +\alpha_p\mb{v}_p.\]
		Letting $\mb{b}=\alpha_{m+1}\mb{v}_{m+1}+\ldots+\alpha_{p}\mb{v}_p$, we have that $\mb{a}-\mb{b} \in \Span(V)$ by construction, which implies that:
		\[\|\mb{b}\|=\|\alpha_{m+1}\mb{v}_{m+1}+\ldots+\alpha_{p}\mb{v}_p\|\geq(p-m)\sqrt{\epsilon}.\]
		This implies that there exists a $\alpha_{j_0}$, $j_0 \in \{m+1,\cdots,p\}$ such that $\|\alpha_{j_0}\|\geq \sqrt{\epsilon}$. Now,
		\[\left\|\sum_{j=1}^p a_i\mb{X}_j\right\|=\|\bm{X}\mb{a}\|=\|\alpha_1 \bm{X}\mb{v}_1+\cdots +\alpha_p \bm{X} \mb{v}_p\|<\epsilon.\]
		We have
\begin{align*}
\epsilon^2&>\mb{a}^T\bm{X}^T\bm{X}\mb{a}\\&=\left(\alpha_1 \bm{X}\mb{v}_1+\cdots +\alpha_p \bm{X} \mb{v}_p\right)^T\left(\alpha_1 \bm{X}\mb{v}_1+\cdots +\alpha_p \bm{X} \mb{v}_p\right)\\&=\alpha_1^2\lambda_1+\cdots +\alpha_p^2\lambda_p\\&\geq \alpha_{j_0}^2\lambda_{j_0}.
\end{align*}
Since  $|\alpha_{j_0}|\geq\sqrt{\epsilon}$, and $\lambda_{j_0}\geq\epsilon$, we have that $\epsilon^2>\alpha_{j_0}^2\lambda_{j_0}\geq \epsilon^2$, a contradiction.
		
		\item[(b)] If $m=p$, then  $\mb{a}=\sum_{j=1}^p a_j \mb{v}_j$. Note $\|\mb{a}\|^2=\sum_{j=1}^p a_j ^2$,  since
		$\mb{v}_j$ are orthonormal.
		Hence, for  $\|\mb{a}\|=1$
		$$ \begin{array}{rcl}
		\|\bm{X}\mb{a}\|^2 & = & \mb{a}^T\bm{X}^T \bm{X} \mb{a} \\
		                      &= &  \displaystyle \left( \sum_{j=1}^p a_j \mb{v}_j \right)^T  \bm{X}^T \bm{X}   \left(\sum_{j=1}^p a_j \mb{v}_j\right) \\
		                      &=&  \displaystyle \left(  \sum_{j=1}^p a_j \mb{v}_j\right)^T      \left( \sum_{j=1}^p a_j  \lambda_j \mb{v}_j\right)  \\
		                      &=&  \displaystyle  \sum_{j=1}^p \lambda_j a_j^2 \\
		                      & < & \epsilon \|\mb{a}\|^2 =\epsilon .
		                      \end{array}$$
		                      leading to $\|\bm{X}\mb{a}\|< \sqrt{\epsilon}.$
		 We assume $m<p$.
		 We  write $\mb{a}$ as:
		\[\mb{a}=\alpha_1\mb{v}_1+\ldots +\alpha_p\mb{v}_p\]
		and observe that:
		\begin{equation}
		\label{E6}
		\min_{\mb{u} \in \Span(V)} \|\mb{a}-\mb{u}\|=\|\alpha_{m+1}\mb{v}_{m+1}+\ldots+\alpha_p\mb{v}_p \|.
		\end{equation}
		
		Since by assumption there exists a $\mb{b}$ with $\|\mb{b}\|<\sqrt{\epsilon}$ and $\mb{a}-\mb{b}\in \Span(V)$,  the vector $\mb{a}-\mb{b}$ is a feasible solution to problem (\ref{E6}), and thus taking $\mb{u}=\mb{a}-\mb{b}$ we have:
		\[\|\mb{a}-(\mb{a}-\mb{b})\|=\|\mb{b}\|\geq \|\alpha_{m+1}\mb{v}_{m+1}+\ldots+\alpha_p\mb{v}_p\|, \]
		leading to:
		\[\|\alpha_{m+1}\mb{v}_{m+1}+\ldots+\alpha_{p}\mb{v}_p\| <\sqrt{\epsilon}.\]
		Since 
		$$\|\alpha_{m+1}\mb{v}_{m+1}+\ldots+\alpha_{p}\mb{v}_p\|^2=\sum_{j=m+1}^{p} \alpha_j^2\ <  \epsilon, $$ Therefore, we have  $|\alpha_j|<\sqrt{\epsilon}$ for all $j=m+1,\ldots,p$.
Thus, we have:
	\begin{eqnarray*}
	\mb{a}^T\bm{X}^T\bm{X}\mb{a}&= & \left(\alpha_1 \bm{X}\mb{v}_1+\cdots +\alpha_p \bm{X} \mb{v}_p\right)^T\left(\alpha_1 \bm{X}\mb{v}_1+\cdots +\alpha_p \bm{X} \mb{v}_p\right)\\
	&=&\alpha_1^2\lambda_1+\ldots +\alpha_p^2\lambda_p\\
	&=& \left(\alpha_1^2\lambda_1+\ldots+\alpha_m^2\lambda_m\right)+\left(\alpha_{m+1}^2\lambda_{m+1}+\ldots+\alpha_p^2\lambda_p\right)\\
	&\leq&  \epsilon \cdot \sum_{j=1}^m \alpha_j^2 +\epsilon \cdot \sum_{j=m+1}^p \lambda_j\\
	&\leq & (1+\lambda_{m+1}+\ldots+\lambda_p)\epsilon,
	\end{eqnarray*}
	leading to $\|\bm{X}\mb{a}\|\leq \sqrt{(1+\lambda_{m+1}+\cdots+\lambda_p)\epsilon}$ as required.
	\end{itemize}
\end{proof}
Theorem \ref{theo2} implies  that if we are able to describe $\Span(V)$, then we would be able to identify multicollinear relationships  $\mb{a}$  that exist in the design matrix $\bm{X}$, as   Theorem \ref{theo2}(b) implies that every vector within $\sqrt{\epsilon}$ distance away from $\Span(V)$ represents a $O(\sqrt{\epsilon})$ multicollinear relationship. 

\subsection{Identifying Multicollinear Relations}
For $\dim(V)=r$, we have $r-1$ linearly independent multicollinear relationships. There are infinite number of ways the basis of the   $r-1$ multicollinear relationships could be constructed, and different ways of constructing such  bases lead to different constraints.  

For example, assume that we have six variables $x_1,x_2,x_3,x_4,x_5,x_6$, and we know that $x_1+x_2=x_3$ and $x_4+x_5=x_6$. 
Letting $\mb{a}_1=(1,1,-1,0,0,0)^T$ and $\mb{a}=(0,0,0,1,1,-1)^T$, we have
$V=\Span (\mb{a}_1, \mb{a}).$
Using Theorem \ref{theo2} and ignoring $\mb{b}$ as $\|\mb{b}\|=O(\sqrt{\epsilon})$, we can identify the two multicollinear relationships as $\mb{a}_1$ and $\mb{a}$. Then, we add the  constraints
\[z_1+z_2+z_3 \leq 2,  \qquad  \qquad z_4+z_5+z_6\leq 2\]
to  Model (\ref{E0}).
However,  there are alternative  ways  to characterize $V$ in terms of two linearly independent vectors. 
Letting $\overline{\mb{a}}_1=(1,1,-1,1,1,-1)^T$ and $\overline{\mb{a}}=(1,1,-1,-1,-1,1)^T$, then
$V$ is also $V=\Span (\overline{\mb{a}}_1, \overline{\mb{a}}).$
Given this representation of $V$ we   would impose the constraints
\[z_1+z_2+z_3+z_4+z_5+z_6\leq4\]
to  Model (\ref{E0}).
Note that the two sets of constraints are not equivalent. 

It is therefore important to   identify the characterization of $V$ that leads to the most stringent constraints to prevent multicollinearity. 
Towards this objective and ignoring the vector $\mb{b}$ in Theorem \ref{theo2} , we introduce the idea of identifying a   vector $\mb{a} \in \Span(V)$  that has minimum support. 
We first compute the set $V=\{\mb{v}_1,\ldots,\mb{v}_m\}$ of orthonormal eigenvectors with corresponding eigenvalues less than $\epsilon$. According to Theorem \ref{theo2}, all multicollinear relationships (up to a perturbation of $\epsilon$) are included in this space. Now, we want  to  find a vector $\mb{a}\in \Span(V) $  of minimum support. This is computed as follows: 

\begin{eqnarray}
\min & \displaystyle \sum_{j=1}^m z_j \label{E7} \\
{\rm subject~to} & \displaystyle \mb{a}= \sum_{i=1}^m  \theta_i \mb{v}_i \nonumber \\
                          & |a_j| \leq M \cdot z_j, \quad j=1,\ldots,m    \label{sos} \\
                          & \displaystyle \sum_{i=1}^m  \left |\theta_i\right | \geq \delta  \nonumber \\
                          & z_j \in \{0,1\},~~j=1,\ldots, m,  \nonumber
\end{eqnarray}                          
Note that (\ref{sos}) can be modeled as Special Ordered Sets (SOS) of type 1, which does not need an explicit value of $M$. In the experiments, however, we utilize the big $M$ formulation. We provide the following procedure for determining $M$.
We  reformulate (\ref{sos}) to read as:
\begin{equation}
\left\|\sum_{i=1}^m \theta_i\bm{v}_i\right\|_1\leq M \sum_{i=1}^m z_j\leq Mm.\label{Mineq1}
\end{equation} 
Using the fact that $\bm{v}_i$'s are orthonormal, we have:
\begin{equation}
\left\|\sum_{i=1}^m \theta_i\bm{v}_i\right\|_1\leq \sqrt{m}\|\bm{\theta}\|_2.
\label{Mineq2}
\end{equation}
Taking $\|\bm{\theta}\|_2=1$,  we can select $\displaystyle M=\frac{\sqrt{m}}{m}=\frac{1}{\sqrt{m}}$. We note this is the tightest possible $M$ with equality at $\bm{v}_i=\bm{e}_i$ and $\bm{\theta}=(\frac{1}{\sqrt{m}},\cdots,\frac{1}{\sqrt{m}})$.

Here $\delta$ is a positive constant that ensures that $\mb{a}\neq 0.$
Once the vector $\mb{a}$ has been identified, we add the constraint 
\begin{equation}
\sum_{i\in \supp(\mb{a})} z_i \leq | \supp(\mb{a})|  -1
\label{E8}
\end{equation}
to Problem (\ref{E0}).
To continue the process of identifying new linearly independent multicollinear relationships, 
we add Eq. (\ref{E8}) to Problem (\ref{E7}), resolve the  problem to  identify a new multicollinear relationship, add the corresponding constraint (\ref{E8}) to 
(\ref{E0}). We continue solving Problem (\ref{E7}) until the problem becomes infeasible, which means that we identified  all  linearly independent multicollinear relationships. 
Algorithm \ref{alg1}    determines all multicollinear relationships.
\begin{algorithm}
	\begin{algorithmic}[1]
		\Procedure{IterativeMC}{$E$}\Comment{$E$, the set of $\bm{X}^T\bm{X}$ eigenvectors with eigenvalues $<\epsilon$}
		\State $S \gets \emptyset$ \Comment{Initialize Output}
		\State $i \gets 0 $ \Comment{Initialize count of found multicollinear relationships}
		\While{$i<|E|$}\Comment{We cannot find more if $i=|E|-1$}
		\State $\mb{a}_0$ = Solution of (\ref{E7})
		\If{$\mb{v}_0 \neq \emptyset$}
			\State $S \gets S \cup \mb{v}_0$
			\State Add  the constraint $\sum_{i \in \supp (\mb{a}_0)} z_i \leq |\supp ( \mb{a}_0)  |-1$ to (\ref{E7})
			\State $i \gets i +1$
		\Else \Comment{If (\ref{E7}) is Infeasible}
			\State Exit Loop
		\EndIf
		\EndWhile
		\State \textbf{return} $S$\Comment{Return the characterization of $V$}
		\EndProcedure
	\end{algorithmic}
	\caption{Iterative MIO for finding  all linearly independent multicollinear relationships.}
	\label{alg1}
\end{algorithm}

\newpage
\subsection{Computational Results}
In this section, we use synthetic data to evaluate the performance of Algorithm \ref{alg1}.

We model the design matrix $\bm{X} \in \md{R}^{n\times p}$ such that $X_{ij} \sim N(0,1)$ independently for each $i \in \{1,\cdots, n\}$, $j \in \{1,\cdots, p\}$. Then we randomly select certain number of columns to be replaced by linear combinations of other columns $\sum_{ij\in S} \gamma_j \bm{X}_j$.  
The  parameters $\gamma_i$ are selected randomly from the uniform distribution $[-10,10]$, and we control $S$ as follows:
\begin{enumerate}
	\item We first determine the number $q$ of variables we want to involve in this multicollinear relationship.  
	\item We randomly select $q$ numbers from $\{1,\ldots,p\}$ without replacement, and denote that set $S$.
\end{enumerate}
We  add noise $\tilde{\bm{X}}$ according to the distribution indicated   in Table \ref{tab3}, and evaluate the performance of Algorithm  \ref{alg1} on $\bm{X} + \tilde{\bm{X}}$.

Algorithm \ref{alg1} performance is evaluated on the accuracy and the false positive rate of the multicollinear relationships found, along with the time taken for the algorithm to converge. 

In  Table \ref{tab3}, $MR(q)$ indicates the number of multicollinear relationships involving $q$ variables that have been introduced into the data. For $MR(4+)
$, we randomly selected a number within $\{5,$ $6,$  $7,$ $8,$ $9,$ $10\}$ to be the  number of variables involved in the multicollinear relationship. We created 10 random instances and report the average statistics across those 10 instances. 
\begin{table}[H]
\centering 
	\begin{tabular}{|l|l|l|l|l|l|l|l|l|}
		\hline $\bm{n}$ & $\bm{p}$ & $\bm{MR(3)}$  & $\bm{MR(4)}$ &  $\bm{MR(4+)}$ & \textbf{Noise} &  \textbf{ACC} & \textbf{FPR} &\textbf{Time}\\\hline
		1000 & 100 & 3 & 1 & 1 & $N(0,0.01)$ & $100\%$& $0\%$& 0.27s\\\hline
		1000 &   500 &3 & 1 & 1& $N(0,0.01)$ & $100\%$&$0\%$& 2.37s\\\hline
		1000 &  1000 & 3 & 1 & 1& $N(0,0.01)$ & $100\%$& $5\%$ &20.23s \\\hline
		1000 &   500 & 5 & 3 & 2& $N(0,0.01)$ & $100\%$& $0\%$& 33.40s\\\hline
		1000 &  1000 &5 & 3 & 2& $N(0,0.01)$ & $100\%$& $24\%$& 5940.56s\\\hline
		1000 &  500 & 3 &1  & 1& $N(0,0.03)$ & $100\%$& $0\%$& 2.29s\\\hline
		1000 & 1000&3  &1 &1 & $N(0,0.03)$ & $100\%$& $11\%$& 32.17s\\\hline
	\end{tabular}
	\caption{Performance of Algorithm \ref{alg1} for  multicollinearity detection.}
	\label{tab3}
\end{table}
Table \ref{tab3}  shows that Algorithm \ref{alg1}   scales up to $n,p$ in thousands and could detect multicollinearity with high accuracy and low false positive rates. 
\section{Holistic Linear Regression Framework Evaluation}
In this section, we combine the results of the previous two sections with the framework introduced in   \cite{SLR1}   on five  different datasets randomly selected from the UCI Machine Learning Repository (\cite{UCIData}). We refer to our framework below as \textit{Holistic}. The whole framework is:
\begin{align*}
    &\min \quad (\ref{E0})\\
    &\text{subject to } (\ref{E1})-(\ref{E4}), (\ref{t-new}),(\ref{t-new1}),(\ref{E8})
\end{align*}
We select $\delta =10^{-6}$ for the multicollinearity detection. We compare our formulation with the MISDO$_{NE}$ formulation ($\kappa = 100$) as denoted in \cite{tamura2017} and the full MIQO formulation (ignoring the alternative solution procedure) by \cite{chung2017}. We note here that although all of the algorithms implement subset selection and their objectives are similar (with the exception of Holistic regression having a $l1$ regularization term), the algorithms differ in their formulation of the constraints. A brief table comparing the relevant constraints is presented below:
\begin{table}[H]
\centering
\resizebox{\textwidth}{!}{
\begin{tabular}{|c|ccc|}
		\hline \textbf{Constraint Type} & \textbf{Holistic}  & \textbf{\cite{tamura2017} }  & \textbf{\cite{chung2017}} \\\hline
		Subset Selection & Big-$M$ & Condition Number-based & Big-$M$\\
	 Significance& Normality-based & None & $t$-test based\\
		Multicollinearity& Explicit & Condition Number-based & None\\
		Residuals & None & None & Absolute \& Breusch-Pagan test\\\hline
\end{tabular}
}

	\caption{Comparison of Constraints within the Holistic, \cite{tamura2017}, and \cite{chung2017} frameworks}
\label{tab4}
\end{table}

We see that compared to the Holistic framework, Tamura does not have an explicit significance constraint (though the condition number constraint to some extent helps select significant variables) and Chung's framework does not have an explicit multicollinearity constraint, replacing it with a residual constraint.

We also used Lasso (\cite{Lasso}), and the framework in \cite{SLR1} (which we denote \textit{Bootstrap}) as baselines. Samples were randomized and we utilized a $60\%/20\%/20\%$ for training, validation, and testing, where the validation set was used for tuning of hyperparameters. This includes the sparsity parameter $k$ in the Holistic, Bootstrap, and Chung's framework (there is no explicit sparsity parameter in Tamura and Lasso) and $\Gamma$ in Lasso and the Holistic framework. We utilized a computer with a i7-5820k 6-core CPU and 16GB of DRAM for all our experiments. Julia 1.0 along with Gurobi 8.0 was used for the Holistic, Chung, Lasso, and the Bootstrap frameworks. Tamura's framework is implemented using SCIP 6.0.0 along with SCIP-SDP 3.1.1 in accordance with the original paper and as Gurobi is unable to handle the semidefinite constraints in the formulation. The results are then compared across the  following dimensions:
\begin{itemize}
	\item \textbf{Sparsity} ($k$) - Number of non-zero variables in the final selected model.
	\item \textbf{Regression Loss} (Loss) - Mean squared error on the test set.
	\item \textbf{Significance} - Percentage of non-zero coefficients in the model that are significant on the $5\%$ level using bootstrap to evaluate.
	\item \textbf{Time} ($T$) - Total time used by the model. Time spent detecting multicollinearity for the Holistic model is shown in brackets.
	\item \textbf{Multicollinearity Accuracy} (MA) - Let $V$ be the set as defined in Theorem \ref{theo2} and $V_{\bm{z}}$ be the corresponding set in the final model using $\epsilon=10^{-2}$. Then we calculate $100\% \times \left(1-\frac{\dim(V_{\bm{z}})}{\dim(V)}\right)$, the percentage of multicollinearity relations "avoided" in the final model. 
		
\end{itemize} 
\begin{table}[H]
\centering
\resizebox{!}{180pt}{%
\begin{tabular}{|llll|lllll|}
		\hline \textbf{Dataset} & \textbf{$\bm{n}$ }  & \textbf{$\bm{p}$ }  & \textbf{Method} &  $\bm{k}$ & \textbf{Loss} &  \textbf{Sign.} &$\bm{T}$ &  \textbf{MA} \\\hline
		Airfoil & 1502 & 5 & \textit{Holistic} & 3 & 558 & $100\%$ & 50s & -\\
		 & & & \textit{Tamura} & 4 & 562 & $75\%$ & 19s & -\\
		 & & & \textit{Chung} & 3 & 558 & $100\%$ & 107s & -\\
		 & & & \textit{Lasso} & 4 & 570 & $75\%$ & 7s &  -\\
		 & & & \textit{Bootstrap} & 4 & 564 & $75\%$ & 39870s & -\\\hline
		Cancer & 568 & 29 & \textit{Holistic} & 7 & 1.71 & $100\%$ & 54s& -\\
		 & & & \textit{Tamura} & 11 & 1.90 & $63\%$ & 410s & -\\
		 & & & \textit{Chung} & 7  & 1.71 & $100\%$ & 310s & -\\
		 & & & \textit{Lasso} & 23 & 0.72 & $31\%$ & 10s &  -\\
		 & & & \textit{Bootstrap} & 12 & 2.22 & $60\%$ & 60000s & -\\\hline	
		Parkinsons & 5875 & 16 & \textit{Holistic} & 1 & 533 & $100\%$ & 403s (15.2s) & $100\%$\\
		 & & & \textit{Tamura} & 1 & 533 & $100\%$ & $60000s$ & $100\%$\\
		 & & & \textit{Chung} & 3 & 549 & $100\%$ & 876s & $50\%$\\
         & & & \textit{Lasso} & 3 & 522 & $33\%$ & 14s &  $50\%$\\
		 & & & \textit{Bootstrap} & 3 & 571 & $33\%$ & 60000s & $50\%$\\\hline	
		Air Quality & 9358 & 12 & \textit{Holistic} & 4 & 89.2 & $100\%$ & 380s & -\\
		 & & & \textit{Tamura} & N/A & N/A & N/A & N/A & -\\
		 & & & \textit{Chung} & 6 & 96.1 & $100\%$ & 770s & -\\
	         & & & \textit{Lasso} & 9 & 83.7 & $33\%$ & 11s &  -\\
		 & & & \textit{Bootstrap} & 5 & 89.6 & $80\%$ & 58146s & -\\\hline	
		Crime & 2215 & 125 & \textit{Holistic} & 9 & 180 & $100\%$ & 725s (120.3s) & $100\%$\\
		 & & & \textit{Tamura} & N/A & N/A & N/A & N/A & N/A\\
		 & & & \textit{Chung} & 11 & 207 & $100\%$ & 1506s & $40\%$\\
	         & & & \textit{Lasso} & 19 & 172 & $47\%$ & 21s & $20\%$\\
		 & & & \textit{Bootstrap} & 12 & 195 & $74\%$ & 60000s & $20\%$\\\hline	
\end{tabular}
}
	\caption{Comparison of the holistic framework with  Lasso for five real world  data  sets. - means that there are no true multicollinear relationships  in the data. N/A means the algorithm did not return a feasible solution within 60000s.
}
\label{tab4}
\end{table}

The results in Table \ref{tab4} show that in real data situations, the entire framework could reasonably scale up to $~10,000$ in $n$ and at least $100$ in $p$, while both the MISDO and the MIQO framework scaled slower. In particular, the MISDO formulation failed to return a feasible solution for the largest datasets, and encountered numerical issues in the process (which was also identified in the original paper). The MISDO formulation also does not consider significance constraints, which was reflected in that some of the variables it selected were insignificant. The MIQO formulation scaled better compared to the MISDO formulation, but was still over $2$x slower than holistic regression in all datasets. Compared with the holistic formulation which only has one set of lazy constraints based on significance, Chung's MIQO formulation has two (significance and residual plots), and we conjecture such additional lazy constriants make it easier for the incumbent solution to be rejected and causes more lazy constraints to be enforced, slowing the runtime. Furthermore, the MIQO formulation does not explicitly model multicollinearity, and this meant the final model in the Parkinsons and the Crime dataset did not avoid more than $50\%$ of multicollinear relationships. In comparison, the holistic regression successfully detected $100\%$ all multicollinear relationships within the data and avoided choosing all variables within that relationship in the final result.We further conjecture that such explicit modeling of significance and multicollinearity constraints is also why the holistic framework usually selects the smallest number of variables, as it needs to satisfy more constraints on subset selection $z$.

Compared to the Lasso baseline, the holistic framework achieved comparable loss with Lasso among most tasks, while using many fewer variables to do so (usually less than half), and all of the selected variables from the framework are significant at the $5\%$ level. The original bootstrap method proposed in \cite{SLR1} quickly timed out as $p$ increased, resulting in suboptimal performance. 

The computational results suggest  that the proposed   holistic  linear regression  algorithm greatly increases its scalability in detecting significance and avoiding multicollinearity. Using both real and synthetic data, we have  demonstrated that the approach produces high quality linear regression models in realistic timelines.
\bibliographystyle{ieeetr}
\nocite{*}
\bibliography{PMCite}
\end{document}